# Two-step Domain Adaptation for Mitosis Cell Detection in Histopathology Images

Ramin Nateghi[1*], Fattaneh Pourakpour[2]
[1]Electrical and Electronics Engineering Department, Shiraz University of Technology, Shiraz, Iran
Iranian Brain Mapping Lab, National Brain Mapping Laboratory, Tehran, Iran

[1*]r.nateghi@sutech.ac.ir, [2] pourakpour@nbml.ir

## ABSTRACT

We propose a two-step domain shift-invariant mitosis cell detection method based on Faster RCNN and a convolutional neural network (CNN). We generate various domain-shifted versions of existing histopathology images using a stain augmentation technique, enabling our method to effectively learn various stain domains and achieve better generalization. The performance of our method is evaluated on the preliminary test data set of the MIDOG-2021 challenge. The experimental results demonstrate that the proposed mitosis detection method can achieve promising performance for domain-shifted histopathology images.

## 1. INTRODUCTION

The number of mitosis cells is one of the critical features in Nottingham Grading systems [1], which is wieldy used for breast cancer grading. Manual mitosis cell counting is a time-consuming task in which a pathologist analyzes the entire tissue. In recent decades, with the advent of whole slide imaging scanners, the entire tissue can be digitized as multiple high-resolution images, encouraging us to develop computerized methods for mitosis cell detection. One of the significant difficulties in mitosis cell detection is the scanner variability and stain variations in tissue [2], which is often driven by differences in staining conditions and tissue preparation and using various scanners. This problem would adversely affect the mitosis cell detection performance, especially when the training and testing data don't come from the same domain distribution. This situation is known as the domain shift problem in the literature [3]. To address this problem, several approaches have been proposed in the literature [4]. Stain normalization is one of the approaches that can be used for domain shift adaptation [5], which is often used as preprocessing before training the network. The stain normalization methods change the color appearance of a source dataset by using the color characteristics of a specific target image. Despite the stain normalization methods often improves the mitosis detection performance, but they sometimes can make an adverse effect on the performance due to not preserving detailed structural information of the cells for all domain shifted cases. Data augmentation is another popular technique that is used for domain shift adaptation [6]. In recent years, several methods have been quantified domain shift effects on model performance. Some recent solutions are based on deep convolutional neural networks and the adversarial neural networks [7]. Lafarge et al proposed a domain-adversarial neural network for removing the domain information from the model representation [8]. In the next sections, we propose two-step domain adaptation for mitosis cell detection based on Faster RCNN and a convolutional neural network.

## 2. DATASET

The data set used in this study is related to an international competition on Mitosis Domain Generalization (MIDOG) [9]. The MIDOG training dataset consists of 200 breast cancer Whole Slide Images (WSIs) stained with Hematoxylin & Eosin (H&E). The samples were scanned with four different scanners including the Hamamatsu XR NanoZoomer 2.0, the Hamamatsu S360, the Aperio ScanScope CS2, and the Leica GT450. Mitosis cells were annotated by pathologists within the selected region of interest with an area of approximately 2mm$^2$. The annotations are only provided for images scanned by three scanners and no annotations were available for the images scanned with Leica GT450. The preliminary test set which is used for performance evaluation consists of 20 images scanned for different scanners.

## 3. METHOD

Our method consists of two steps: the preliminary mitosis cell detection by the Faster RCNN model and final mitosis cell detection with a CNN. Fig. 1, represents the block diagram of our mitosis cell detection method. In the first step, we use Faster RCNN to detect mitosis cell candidates. To overcome the domain shifting problem, we used a stain augmentation tool called StainTools for domain generalization in which we randomly generate ten new histopathology images with a shifted stain appearance from each original training image. This augmentation technique not only helps us to extend the training dataset but can also effectively improve Faster RCNN model generalization. Fig. 2 represents some generated augmented images for a sample region of histopathology image. Having false-positive results is a most challenging problem for mitosis detection. Therefore, in the second step, the detected mitosis cells are used to train a CNN to perform finer mitosis detection.

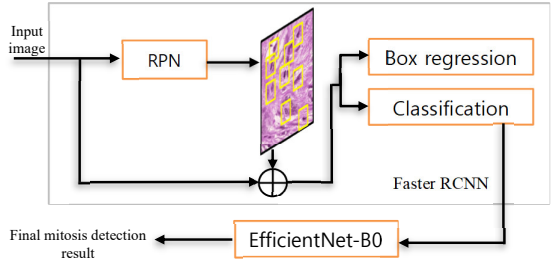

Fig. 1: The proposed mitosis cell detection method

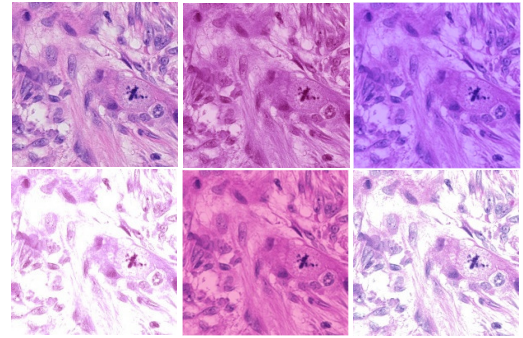

Fig. 2: The stain augmentation with StainTools

### 3.1 Training parameters:

We only used the images scanned by Hamamatsu XR NanoZoomer, Hamamatsu S360, the Aperio ScanScope CS2 scanners for the training, since the annotations have been only provided for them. For network training, instead of splitting our training dataset into two training and validation subsets, we used the k-fold cross-validation technique (k=4) as a preventative technique against overfitting. Using this technique, the dataset is randomly divided into four different subsets. Because the images are large in size, the images of each subset are split into small patches with the size of 1536×2048 (the padding is done if needed). In the next step, the mentioned augmentation technique is used to expand each subset in order to improve the performance and the model generalization. Then we trained four Faster RCNN models using the four augmented subsets. During the training of each model, one subset is considered as the validation set and the remaining as the training set. For model training, we used a mini-batch size of 4, with a cyclical maximal learning rate of $10^{-4}$ for 40 epochs by considering binary cross-entropy and smooth L1 losses for classification and regression heads respectively. The validation loss is also used for the early stopping and checkpoint (with a patience of ten epochs), helping the models to further avoid overfitting. For combining the results of four trained Faster RCNN models we used Weighted Boxes Fusion (WBF) [10].

After detecting the mitosis cell candidates, the second mitosis cell detection step is performed. All of the false-positive and truly detected mitosis cells at the output of the first step are used to train EfficientNetB0 networks. Four different networks are trained in the second step using the detected cells within the four subsets. Before training the networks, we extended the cell subsets using the proposed augmentation technique for domain generalization. For the training, we used a mini-batch size of 256 and trained the models for 200 epochs with a cyclical maximal learning rate of $10^{-4}$. To avoid overfitting, the early stopping with a patience of fifty epochs is used during training. The binary cross-entropy loss is also considered to train the networks.

### 4. EVALUATION AND RESULTS

We evaluated the performance of the proposed method on the preliminary test set. Table .1 summarizes the performance of our mitosis detection method on the preliminary test set based on three criteria including precision, recall, and F1 score. The precision represents the percentage of the truly detected mitosis cells, while recall expresses the rate of real mitosis cells, and the F1 score is the harmonic mean of precision and recall. We individually evaluated the performance of the first step mitosis detection results to better understand the importance of the multi-stage classification in reducing the false-positive results. Despite using a huge augmented dataset for training, the first mitosis detector achieved an F1 score of 40.28% on the preliminary test set, containing some false positives at the output. Our best result on the preliminary set was obtained when using the proposed two-step mitosis detection method, resulting in an F1 score of 68.95%. The results demonstrated that the second classification step considerably reduced the false positives and significantly improved the mitosis detection performance.

Table 1. Performance of our method on the preliminary test set

| Method | Precision | Recall | F1-score |
|---|---|---|---|
| one-step classification (Faster RCNN) | 26.46 % | 84.33 % | 40.28 % |
| Two-step classification (Faster RCNN+ EfficientNetB0) | 65.41 % | 72.89 % | **68.95 %** |

## 5. CONCLUSION

In this work, we presented a two-step domain shift-invariant mitosis cell detection method based on Faster RCNN and CNN models. We used a stain augmentation technique for domain generalization as well as dataset expansion. Experimental results demonstrated the promising domain generalization of our model for mitosis cell detection.